\documentclass[10pt,twocolumn,letterpaper]{article}

\usepackage{wacv}
\usepackage{times}
\usepackage{epsfig}
\usepackage{graphicx}
\usepackage{amsmath}
\usepackage{amssymb}

\usepackage{framed,multirow}
\usepackage{booktabs, arydshln} 
\usepackage{makecell}

\usepackage{verlab}
\usepackage{framed,multirow}
\usepackage{amssymb}
\usepackage{latexsym}
\usepackage{hyperref}
\usepackage{url}
\usepackage{times}
\usepackage{epsfig}
\usepackage{graphicx}
\usepackage{amsmath}
\usepackage{amssymb}
\usepackage{relsize}
\usepackage{booktabs, arydshln} 
\usepackage[utf8]{inputenc}
\usepackage{mathtools}
\usepackage{makecell}
\usepackage{array}
\usepackage{float}
\usepackage{nicefrac}
\usepackage[dvipsnames]{xcolor}
\usepackage{xparse}
\usepackage{microtype}

\definecolor{newcolor}{rgb}{.8,.349,.1}

\usepackage{verlab}
\usepackage{fancyhdr}
\DeclareMathOperator*{\argmin}{argmin} 

\makeatletter
\def\adl@drawiv#1#2#3{%
	\hskip.5\tabcolsep
	\xleaders#3{#2.5\@tempdimb #1{1}#2.5\@tempdimb}%
	#2\z@ plus1fil minus1fil\relax
	\hskip.5\tabcolsep}
\newcommand{\cdashlinelr}[1]{%
	\noalign{\vskip\aboverulesep
		\global\let\@dashdrawstore\adl@draw
		\global\let\adl@draw\adl@drawiv}
	\cdashline{#1}
	\noalign{\global\let\adl@draw\@dashdrawstore
		\vskip\belowrulesep}}
\makeatother
\NewDocumentCommand{\rot}{O{45} O{1em} m}{\makebox[#2][l]{\rotatebox{#1}{#3}}}%

\wacvfinalcopy 

\def\wacvPaperID{485} 

\ifwacvfinal\fi
\begin{document}

\title{Personalizing Fast-Forward Videos Based on Visual and Textual Features from Social Network
}

\author{Washington L. S. Ramos \hspace{.5cm} Michel M. Silva \hspace{.5cm} Edson R. Araujo \hspace{.5cm} Alan C. Neves\\Erickson R. Nascimento\\
Universidade Federal de Minas Gerais (UFMG), Brazil\\
{\tt\small \{washington.ramos, michelms, edsonroteia, alan.neves, erickson\}@dcc.ufmg.br}
}

\maketitle
\thispagestyle{fancy}
\fancyhf{}
\chead{{To appear in Proceedings of the IEEE Winter Conference on Applications of Computer Vision (WACV) 2020 \\ The final publication will be available soon.}}

\begin{abstract}
	The growth of Social Networks has fueled the habit of people logging their day-to-day activities, and long First-Person Videos~(FPVs) are one of the main tools in this new habit. Semantic-aware fast-forward methods are able to decrease the watch time and select meaningful moments, which is key to increase the chances of these videos being watched. However, these methods can not handle semantics in terms of personalization. In this work, we present a new approach to automatically creating personalized fast-forward videos for FPVs. Our approach explores the availability of text-centric data from the user's social networks such as status updates to infer her/his topics of interest and assigns scores to the input frames according to her/his preferences. Extensive experiments are conducted on three different datasets with simulated and real-world users as input, achieving an average F\textsubscript{1} score of up to $12.8$ percentage points higher than the best competitors. We also present a user study to demonstrate the effectiveness of our method.
\end{abstract}

\section{Introduction}
\label{sec:introduction}

%
The past decade has witnessed an explosion of tools for Internet users to share their interests and day-to-day activities with each other. The most representative tools are social multimedia services such as Twitter, Facebook, and YouTube, where the users upload and describe relevant information about themselves. Most recently, wearable cameras have emerged as a promising and effective tool for people to document their lives. The high storage capacity and long battery life of these devices foment continuous recording, resulting in massive streams of raw footage further uploaded to the online social media. While it is easy to produce and store lengthy First-Person Videos (FPVs), they are unlikely to be revisited, even if they contain meaningful moments for the recorders and their followers.

%
Although video summarization techniques can provide a summary with meaningful moments, such approaches are limited in presenting only fragments of the original video, causing a temporal gap in the storyline. There is a growing body of research on hyperlapse methods~\cite{Karpenko2014, Kopf2014, Poleg2015, Joshi2015, Halperin2017, Rani2018, WangM2018}. These methods create a continuous flow of the timeline by selecting a subset of frames regarding the stability of the inter-frame transitions and the final video length. As a result, the output video contains a fewer number of jerky scene transitions, and their frames are temporally connected.

\begin{figure}[t!]
	\centering
	\includegraphics[width=.99\columnwidth]{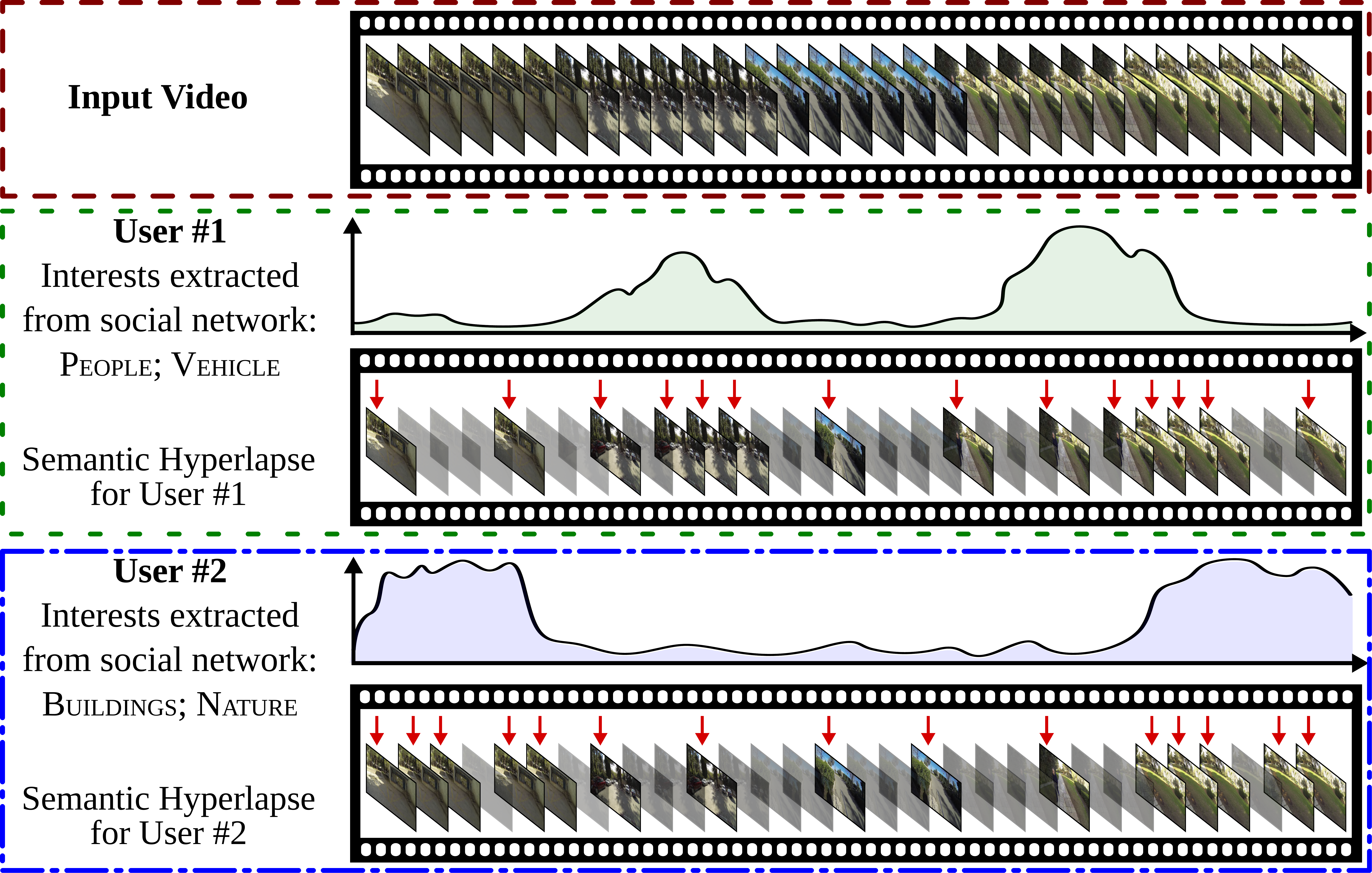}
	\caption{Given a first-person video and user's social network as inputs, our method infers the preferences of the user, calculates a per-frame interestingness score, and selects the frames (red arrows) to create a personalized hyperlapse emphasizing interests.}
	\label{fig:introduction}
\end{figure}

%
Most recent hyperlapse approaches~\cite{Ramos2016, Silva2016, Higuchi2017, Lai2017, Silva2018jvci, Silva2018cvpr} sample the input frames according to the semantic load to emphasize the relevant video segments. A major obstacle faced by these techniques is the encoding of the semantic information. They typically use a predefined set of objects -- \eg, faces and pedestrians~\cite{Ramos2016, Silva2016}, or classes from the PASCAL-Context dataset~\cite{Lai2017}. Despite remarkable advances in using predefined objects of interest, this approach may not successfully be applied to FPVs, since they are shared in social networks where there is a wide range of users and a variety of preferences. People conceivably have distinct preferences in retaining some moments rather than others~\cite{Varini2017, Sharghi2017}.

%
Social Networks have become an underlying channel for people to interact and expose their feelings, emotions, attitudes, and opinions. Despite the broad usage of images and videos, texts are primarily used by the users to describe their preference over specific topics. Motivated by the success of joint vision-language models~\cite{Fang2015, Venugopalan2015, Sharghi2016, Kottur2016, Vinyals2017, Karpathy2017, Donahue2017, Plummer2017, Sharghi2017, Dong2018}, in this paper, we explore the text-centric data from the users' social networks to create personalized hyperlapse videos. We propose to build a unique representation space that encodes video frames and preferences of social network users. Each dimension in the created space defines a topic of interest represented by a set of similar concepts. A user is represented by the frequency of her/his preferred concepts, while each frame is represented by a composition of visual and textual features of its concepts. We compute the similarity between these representations to obtain interestingness scores over the whole video and define the segments of higher relevance. The emphasis on the relevant segments is achieved by reducing the playback rate of such segments in the hyperlapse video. Fig.~\ref{fig:introduction} depicts an example of two personalized hyperlapse videos for users with different topics of interest using the same input video in our method. The interestingness score curve, along with a threshold defines the segments with higher relevance for the different users.

We demonstrate the effectiveness of our approach on three FPV datasets using simulated and real users' data from Twitter. Regarding personalization, our approach presents an F\textsubscript{1} score of up to $ 12.8 $ percentage points higher than the best competitors without causing visual instability in the output video. Moreover, we conduct a user study to assess our results on personalization and visual smoothness.

%
%
In summary, our contributions are: \textit{i)} a novel approach that personalizes a hyperlapse video emphasizing the relevant segments according to the user's topics of interest inferred from her/his social network profile; \textit{ii)} a model for encoding the user and video frames semantics in the same representation space, capable of leveraging raw concepts to topics. In other words, if in written texts the user says she/he likes birds, our model generalizes birds to nature. Therefore, the hyperlapse video emphasizes segments where nature-related elements (birds, plants, trees, \etc) appear.

\section{Related Work}
\label{sec:related_work}

In the past years, the problem of creating summaries from long FPVs has been extensively studied. In video summarization, the primary goal is to select the meaningful keyframes or video shots from an input video. Some methods allow the user to personalize the final summary based on her/his preferences~\cite{Varini2017, Sharghi2017, Plummer2017}. However, these solutions do not create a pleasant experience for the user to follow the video storyline, since they output skims that are not temporally connected.

Hyperlapse algorithms, instead, tackle the problem of video discontinuity by prioritizing temporal continuity and video smoothness constraints considering a budget for the number of output frames \cite{Karpenko2014, Kopf2014, Joshi2015, Halperin2017, Rani2018, WangM2018}. Most recent approaches focus on adaptive frame selection. A representative method in this category is the work of Joshi~\etal~\cite{Joshi2015}, in which the authors proposed to adaptively select frames subject to speed-up and smoothness restrictions, presenting a state-of-the-art performance in video stability.

Despite the advances in fast-forwarding FPVs, traditional hyperlapse methods accelerate the entire video disregarding the semantic content, turning the exciting moments, usually short clips in a lengthy video, almost imperceptible. Semantic fast-forward techniques~\cite{Ramos2016, Silva2016, Yao2016, Lai2017, Silva2018jvci, Silva2018cvpr, Lan2018} try to avoid losing relevant events by emphasizing the important segments of the input video. These methods usually segment the video according to the semantic content in frames and apply different speed-up rates according to their relevance.

The definition of what is relevant plays a central role in the whole process of semantic fast-forwarding. Some approaches embed the semantic information using a predefined set of objects~\cite{Ramos2016, Silva2016, Silva2018cvpr}. Alternatively, other approaches define semantics based on general preferences. Yao~\etal~\cite{Yao2016} used majority voting over annotations from $3$ individuals to measure the relevance of a segment. Silva~\etal~\cite{Silva2018jvci} proposed the \textit{CoolNet}, a Convolutional Neural Network (CNN) model, to identify images that are similar to frames composing the most enjoyable videos on the YouTube platform. Although semantics can be personalized by adjusting the training data to videos liked by the user, no massive data from a single user is available to train a CNN. Most recently, Lan~\etal~\cite{Lan2018} introduced the FastForwardNet (FFNet), a reinforcement learning-based method that selects the most important frames without processing the entire video. Despite their results, the performance of FFNet in unconstrained FPVs is still unknown. Also, the video smoothness constraint is neglected in their learning process. 

It is worth noting that the relevance of frames in FPVs recorded in unconstrained scenarios is strongly dependent on the watchers' interest, which may have different preferences over the same video~\cite{Varini2017}. Lai~\etal~\cite{Lai2017} proposed a personalized semantics technique which provides a list of objects and asks the users to select their preferences. The drawbacks include requiring the user interaction and the limited number of objects, which are the $60$ classes from the Pascal-Context Dataset~\cite{Mottaghi2014}. Moreover, the method is not designed to work with regular FPVs, but with ${360^\circ}$ videos.

\begin{figure*}[t!]
	\centering
	\includegraphics[width=0.91\linewidth]{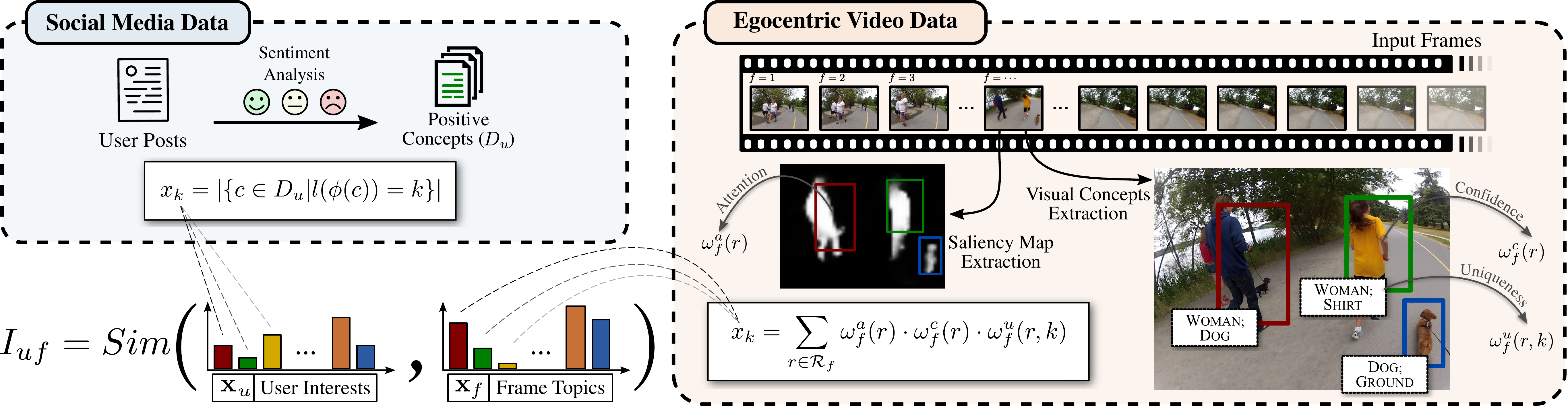}
	\caption{Main steps to compute the Interestingness Score ($ I_{uf} $). We extract concepts from positive posts to compose bins (topics of interest) in the user representation $ \mathbf{x}_u $ (left). For each input frame, we create a vector $ \mathbf{x}_f $ in which the bins are composed of the attention, confidence, and uniqueness weights of each concept in the frames (right). The final score is the similarity between $ \mathbf{x}_u $ and $ \mathbf{x}_f $.}
	\label{fig:frame_scoring}
\end{figure*}

In this paper, we present an approach that correlates the text from the user's social network and frames of the input video to infer the topics of the user's interest and the relevant content to that specific user. Our ultimate goal is to create a fast-forward video emphasizing the segments that are relevant to a specific user, regarding the hyperlapse restrictions of length and visual smoothness.

\section{Methodology}
\label{sec:methodology}
The goal of our approach is to infer the user's preferences from raw input texts in her/his social network and create a personalized hyperlapse video. Formally, let the input video $ {V = \{v_f\}_{f = 1}^F} $ be a sequence of $ F $ frames. We aim at selecting a subset $ {\hat{V} \subset V} $ with the most relevant frames while preserving visual smoothness, temporal consistency, and the speed-up rate $ S $ to achieve the desired number of frames. Our methodology is composed of two major steps: Frame Scoring and Hyperlapse Composition.

\subsection{Frame Scoring}
\label{sec:frame_scoring}

The first step of our methodology identifies and quantifies the amount of semantics in frames according to the users' preferences. As stated by Sharghi~\etal~\cite{Sharghi2017}, concepts can better express the semantic information in terms of what we see in a video. Also, the ability to relate concepts to fragments of videos helps to create meaningful summaries~\cite{Plummer2017}. Therefore, we use concepts to associate frames and users.

\paragraph{Representation Space.} We build a representation space which can be shared between the frames content and the user. In this space, each dimension represents a topic of interest consisting of a set of semantically similar concepts (\eg, guitar, violin, and cello comprising `string instruments'). 

To create such space, we use the distributed word representation framework, \textit{word2vec}~\cite{Mikolov2013}, to learn real-valued vectors lying in a $d$-dimensional embedding space where similar words in a context share a vicinity. Let ${\mathcal{W} = \{\mathbf{w}_i \in \mathbb{R}^d | i = 1, \dots, N\}}$ be the set of such vectors (also known as word embeddings), where $ N $ is the number of vectors. We cluster all $ {\mathbf{w}_i} $ embeddings into $ K $ clusters using the $k$-means algorithm and label each embedding by computing ${l(\mathbf{w}_i) = \arg\min_k \lVert \mathbf{q}_k - \mathbf{w}_i \rVert_2}$, where $ {\mathbf{q}_k \in \mathbb{R}^d} $ is the centroid of the $ k $-th cluster. Because similar concepts are closer in the embedding space, we assume that each cluster defines a topic of interest. Therefore, concepts extracted from the frame's content or user texts can be used to compose a vector $ \mathbf{x} = [x_1, x_2, \cdots, x_K]^\intercal \in \mathbb{R}^{K} $ lying in a new $K$-dimensional representation space, with each dimension $ x_k $ representing a topic of interest. We use human-annotated region descriptions of the Visual Genome (VG) dataset~\cite{Krishna2017} as a corpus for training the \textit{word2vec} since it benefits optimizing the proximity of visually similar concepts in the embedding space. To include words from the social network vocabulary, we initialize the \textit{word2vec} with the parameters of a pre-trained model generated from a corpus of $ {198} $ million of \textit{tweets} (posts on Twitter) and $ {6.7} $ billion of words from general data (Wikipedia, Google News, \etc)~\cite{Li2017}.

We refer to $ \mathbf{x} $ as a \textit{Bag of Topics} (BoT) representation due to similarities with the \textit{Bag of Features} technique~\cite{FeiFei2005}. Note that this approach is different from the one described by Passalis~\etal~\cite{Passalis2018} since it optimizes the distance between visual concepts straightforwardly, preserving the unsupervised characteristic of the whole process.

\paragraph{User Interests.} In social networks, users commonly share their everyday activity through posts and comments, which are undoubtedly high-level cues of preferences and opinions to a specific topic, and Twitter is one of the most popular platforms for this habit. Due to the noisy and complex nature of \textit{tweets}, obtaining topics of interest is a challenging task \cite{Michelson2010, Bhattacharya2014}. Motivated by these aspects, we use the Twitter API to gather user \textit{tweets} and extract the concepts. Nevertheless, it is worth noting that our approach can be expanded to posts on Facebook, comments on YouTube videos or captions of Instagram pictures. An essential step to inferring the user's preferred concepts is to filter the collected posts and use only the positive ones. We extract the nouns from these posts to represent the user's preferred concepts. Let $ {D_u = \{c_j | j = 1 \dots C\}} $ be a document composed of $ {C} $ concepts and $ {\phi : D \to \mathcal{W}} $ be the function that maps a concept $ c $ to a word embedding vector $ \mathbf{w} \in \mathcal{W} $. Therefore, given $ {D_u^k = \{c \in D_u | l(\phi(c)) = k\}} $, we can represent $ D_u $ as the user BoT representation $ {\mathbf{x}_u \in \mathbb{R}^{K}} $, where $ {x_k = |D_u^k|} $. Fig.~\ref{fig:frame_scoring}-left shows the steps to compose $ {\mathbf{x}_u} $.

\begin{figure}[t!]
	\centering
	\includegraphics[width=\columnwidth]{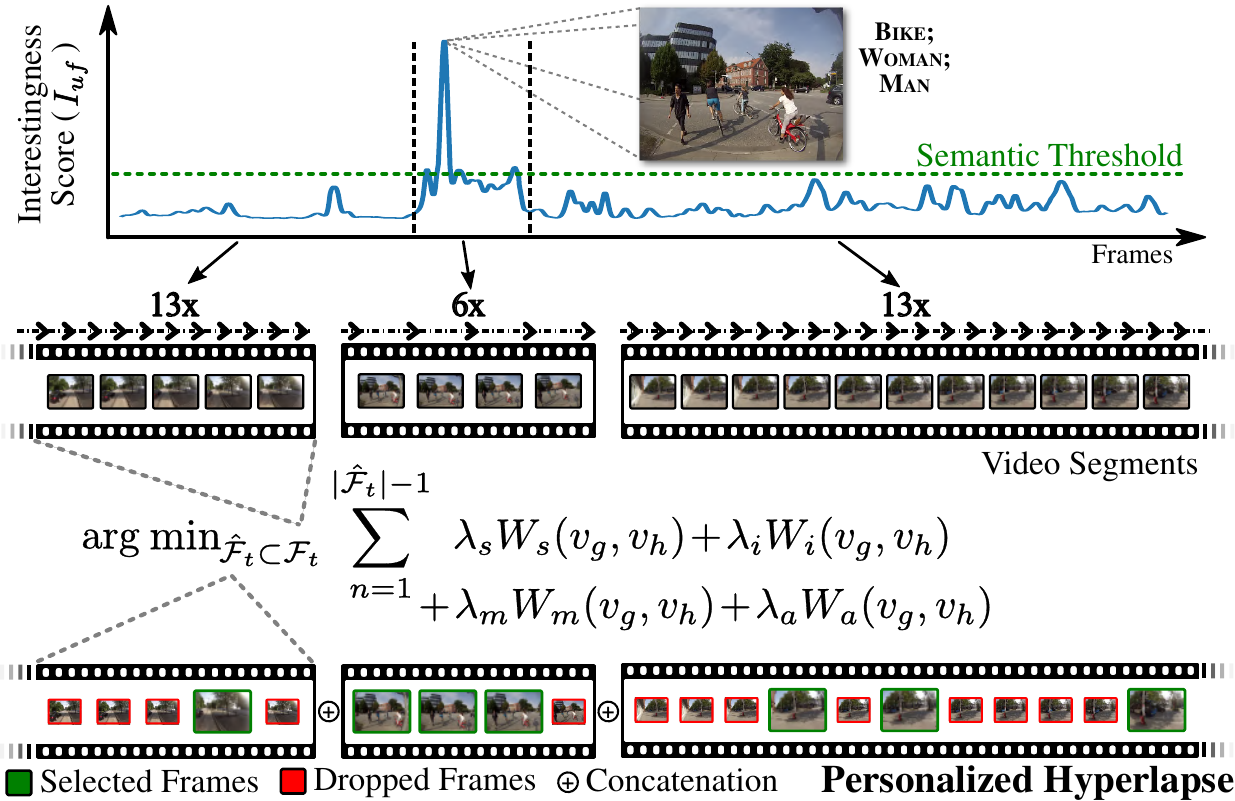}
	\caption{Personalized hyperlapse composition. We first calculate a per-frame interestingness score, then we segment the video into relevant and non-relevant segments, and calculate speed-ups for each type of segment, such that lower speed-ups are assigned to more relevant segments. The sampling process minimizes the costs for semantics, instability, motion, and appearance.}
	\label{fig:frame_sampling}
\end{figure}
\paragraph{Frame Topics.}
To represent a frame $ v_f $, we extract a set $ \mathcal{R}_f $ with $ R $ regions of interest and their respective coordinates, scores, and dense per-region natural language descriptions (a set of sentences $ {D_f = \{s_1, s_2, s_3, \cdots, s_R\}} $) using the DenseCap algorithm~\cite{Johnson2016}. Then, we combine features related to visual and textual cues to assign weights to each region $ r \in \mathcal{R}_f $. We compute the following weights:

\noindent \textit{(i) Attention}. To weight the viewer attention to the region~$ r $, we applied the algorithm of Wang~\etal~\cite{WangW2018}, which uses temporal and motion information to detect salient objects in videos. It produces a probability map $ {P \in [0, 1]^{X \times Y}} $ from $ v_f $, where $ X $ and $ Y $ are the width and height of the frame, respectively. Let $p(x, y)$ be the intensity of a pixel in $ P $ located at $ x $ and $ y $ coordinates, and $ M_r $ be the number of pixels in the region $ r $. The attention weight is computed as:
\begin{equation}
\label{eq:AttentionScore}
\omega_f^a(r) = \frac{1}{M_r} \sum_{x, y \in r} p(x, y).
\end{equation}
\noindent \textit{(ii) Confidence}. The confidence weight, $ \omega_f^c(r) $, is the score assigned to~$ r $~by DenseCap. Higher confidence correspond to more accurate regions, leading to better visualization of the contents within.

\noindent \textit{(iii) Uniqueness}. This weight reflects the importance of the concepts in $ r $ for the whole video story. We handle the video as a collection of documents $ {\mathcal{D} = \{D_f\}_{f=1}^{F}} $ and calculate the log-normalized Term Frequency-Inverse Document Frequency (TF-IDF) of each term in the sentence $ {s_r \in D_f} $ that describes the region $ r $. Thus, the weight is computed as: 
\begin{equation}
\label{eq:UniquenessScore}
\omega_f^u(r, k) = \sum_{c \in C_r} T(c, D_f)[l(\phi(c)) = k],
\end{equation}
\noindent where $ C_r $ is a document composed of the concepts in the sentence $ s_r $, $[\cdot]$ is an indicator function that returns $ {1} $ if the proposition is satisfied and $ {0} $, otherwise, and
\begin{equation}
\label{eq:tf_idf}
T(c, D_f) = ( 1+\log(|\{c \in D_f\}|) ) \cdot \log\Bigg(\frac{\sum_{f = 1}^{F} |D_f|}{|\{c \in D_f\}|} \Bigg).
\end{equation}
The uniqueness score benefits visual concepts that are distinct in the video; therefore, these concepts might attract the viewer interest. Note that, although the inverse document frequency by itself can be used to compute the uniqueness, combining it with the term frequency helps to avoid false positives receiving high scores.

The final weight for the topic $ x_k $ in $ {\mathbf{x}_f \in \mathbb{R}^K} $ is obtained as:
\begin{equation}
\label{eq:topics_frame}
x_k = \sum_{r \in \mathcal{R}_f} \omega_f^a(r) \cdot \omega_f^c(r) \cdot \omega_f^u(r, k).
\end{equation}
We denote $ {\mathbf{x}_f}$ as the BoT representation of the frame $ v_f $. Fig.~\ref{fig:frame_scoring}-right shows the process to compose the vector $ {\mathbf{x}_f} $.

\paragraph{Interestingness Score.} After computing the new semantic representation for both the video frames and user, we can estimate the interest of the user for any given image using a vector similarity metric. In this work, we use the cosine similarity between $\mathbf{x}_u $ and $ \mathbf{x}_f $:
\begin{equation}
\label{eq:interestingness_score}
{I_{uf} = \frac{\mathbf{x}_u^\intercal \mathbf{x}_f} {\lVert\mathbf{x}_u\rVert_2 \lVert\mathbf{x}_f \rVert_2}}.
\end{equation}
\subsection{Hyperlapse Composition}

The frame selection step in our the personalized hyperlapse is presented in Fig.~\ref{fig:frame_sampling}. We compute a per-frame interestingness score to create a profile curve for the input video $ V $ given the texts from user $ u $. Then, we partition $ V $ into $ T $ segments $ {\mathcal{F}_t = \{v_{t,1}, v_{t,2}, \cdots, v_{t,M_t}\}}$, with $ {t = 1, \cdots, T} $ and ${M_t = |F_t|}$~\cite{Silva2016}. A semantic threshold is defined to classify each $ \mathcal{F}_t $ as relevant or not. Segments with higher interestingness scores are classified as relevant, while the others are classified as non-relevant.

In the following, we calculate different speed-up rates for each type of segment such that the relevant segments receive a lower speed-up rate, $ {S_s} $, to maximize the exhibition of relevant concepts. Consequently, the speed-up rate of non-relevant segments, $ {S_{ns}} $, must be higher to keep the overall speed-up $ S $ unchanged. Let $ L_s $ be the overall length of the relevant segments and $ {L_{ns}} $ the overall length of the non-relevant segments, \ie, $ {|V| = L_s + L_{ns}} $. We estimate the final speed-up rates $ {S_s^\ast} $ and $ {S_{ns}^\ast} $ by optimizing:
\begin{equation}
\argmin_{S_{s}, S_{ns}} \Bigg|\frac{|V|}{S} - \Bigg(\frac{L_s}{S_{s}} + \frac{L_{ns}}{S_{ns}}\Bigg)\Bigg| + \lambda_1|S_{ns} - S_{s}| + \lambda_2|S_{s}|.
\label{eq:speedup_rates}
\end{equation}
The terms $ {\lambda_1|S_{ns} - S_{s}|} $ and $ {\lambda_2|S_{s}|} $ ensure $ {S_{s}} $ to be as minimum as possible, limited by difference of speed-ups.

Because the difference between $ S_s^\ast $ and $ S_{ns}^\ast $ may produce abrupt speed-up rate changes, we perform a speed-up rate refinement process~\cite{Silva2018jvci}. We concatenate the relevant segments using them as a new input video. Then, we iterate over the partitioning and speed-up estimation steps using a new target speed-up $ {S = S_s^\ast} $. This process repeats while the new semantic threshold increases by a factor of $ \gamma = 0.2 $. In addition to decreasing the abrupt speed-up changes, this refinement process assigns even lower speed-up rates to segments of higher interest, producing greater emphasis on them.

A primary concern is to satisfy the hyperlapse requirements, \ie, guarantee continuity in the storyline, smoothness in the frames transitions and desired speed-up rate accuracy. Therefore, we model the transitions for any given pair of frames ($ v_g, v_h $) with the following inter-frame costs: (i) the relevance drop cost, which is computed as $ {W_s(v_g, v_h) = \nicefrac{1}{(I_{ug} + I_{uh} + \epsilon)}}$, where $ \epsilon $ prevents the division by $ 0 $ when both frames are completely irrelevant for the user; (ii) the instability cost, $ W_i(v_g, v_h) $, which indicates the average distance of the focus of expansion to the center of the frame~\cite{Sazbon2004}; (iii) the speed of motion cost, $ W_m(v_g, v_h) $, which is computed as the difference of the average magnitude of the optical flow vectors between the pair ($ v_g, v_h $) and the overall average magnitude of the optical flow vectors for every pair of frames temporally distant by a factor of $ S_t^\ast $ (the speed-up estimated for the $ t $-th segment) and; (iv) the appearance cost, $ W_a(v_g, v_h) $, which measures the dissimilarity between $ v_g $ and $ v_h $. We use the Earth Mover's Distance between their color histograms. Halperin~\etal~\cite{Halperin2017} present a more detailed description of how to compute the last three costs. The lower are these costs, the better is the transition between $ v_g $ and $ v_h $. The overall transition cost is the weighted sum of individual cost terms:
\begin{equation}
\label{eq:overall_transition_costs}
\begin{split}
E(v_g,v_h) = \lambda_{s} W_s(v_g,v_h) + \lambda_{i} W_i(v_g,v_h)
\\
+ \lambda_{m} W_m(v_g,v_h) + \lambda_{a} W_a(v_g,v_h).
\end{split}
\end{equation}

For each segment $ \mathcal{F}_t $ we obtain the selected frames $ \mathcal{\hat{F}}_t $ by solving the following minimization problem:
\begin{equation}
\label{eq:frame_selection}
\argmin_{\mathcal{\hat{F}}_t \subset \mathcal{F}_t} \sum_{n = 1}^{|\mathcal{\hat{F}}_t| - 1}
E(\hat{v}_{t,n}, \hat{v}_{t,n+1}),
\end{equation}
\noindent where $ \hat{v}_{t,n} $ is the $ n $-th selected frame in the $ t $-th segment.

To solve Eq.~\ref{eq:frame_selection}, we build a graph for each segment where the nodes represent the frames and edges represent the inter-frame transitions. The weight for the edge connecting a pair of frames ($ v_g $, $ v_h $) is given by Eq.~\ref{eq:overall_transition_costs}. Edges are connected up to a temporal distance of $ \tau = 100 $ frames. The nodes composing the shortest path are the selected frames of the segment. We apply a multiplication factor of $ \left\lceil{\nicefrac{\delta(\hat{v}_{t,n}, \hat{v}_{t,n+1})}{S_t^\ast}}\right\rceil $ to each edge to discourage frame skips larger than $ S_t^\ast $, with $ \delta(v_g, v_h) $ being the temporal distance, in number of frames, between $ v_g $ and $ v_h $~\cite{Ramos2016}.

Finally, we concatenate all the selected frames in each segment (Fig.~\ref{fig:frame_sampling}-bottom) to compose the personalized hyperlapse video $ \hat{V} $. We perform a 2D stabilization to eliminate the remaining jitter in the final video. To this end, we use the fast-forward egocentric video-aware stabilizer proposed by Silva~\etal~\cite{Silva2016}.

\section{Experiments}
\label{sec:experiments}

To evaluate our method, we conducted several experiments using real and simulated users with interests in specific and diverse topics over input videos from different datasets.

\subsection{Experimental Setup} 
\paragraph{Datasets.} We used three datasets in our experiments: the UT Egocentric (UTE) Dataset~\cite{Lee2012}; the Semantic Dataset~\cite{Silva2016}; and EgoSequences, which is composed of videos used to evaluate previous hyperlapse methodologies~\cite{Ramos2016, Halperin2017}. 

The UTE Dataset consists of $ 4 $ first-person videos with $ 3 $-$ 5 $ hours of daily egocentric activities each. Sharghi~\etal~\cite{Sharghi2017} provide human-annotated concepts for this dataset. A binary semantic vector indicates the presence of concepts in each shot of $ 5 $ seconds. The Semantic Dataset is composed of $ 11 $ first-person videos presenting three different activities: biking, driving and walking. EgoSequences is composed of $ 9 $ first-person videos depicting indoor and outdoor activities.

\paragraph{Methods for comparison.} We compared our method against three fast-forwarding approaches: i) \textit{Uniform}, which samples one frame at every $ S $-th frame of the input video, where $ S $ is the required speed-up rate; ii) \textit{Microsoft Hyperlapse} (MSH) \cite{Joshi2015}, which adaptively selects frames from the input video optimizing for a smooth camera motion, as well as the target speed-up and; iii) \textit{Multi-Importance Fast-Forward} (MIFF) \cite{Silva2018jvci} that extracts the semantic information by detecting faces and pedestrians on each frame.

For the sake of a fair comparison, we do not compare with video summarization approaches because of the lack of visual smoothness and speed-up constraints in these techniques. Although some works do include temporal coherence in their design, such constraint does not play a central role in the selection of frames as in hyperlapse works.

\paragraph{Evaluation metrics.} We evaluate the methodologies in terms of personalization, speed-up rate accuracy, and instability of the output videos. To measure personalization, we used the F\textsubscript{1} score, which is the harmonic mean of precision and recall. An output video reaches the best F\textsubscript{1} score if all frames from relevant segments were selected, and the whole video is composed only of relevant frames. Regarding the accuracy of the speed-up rate, we measure the deviation to the required rate by calculating $ | S - \hat{S} | $, where $ {\hat{S} = \nicefrac{|V|}{|\hat{V}|}} $ is the final speed-up rate and $ {\hat{V} = \bigcup_{t=1}^{T}} \mathcal{\hat{F}}_t$. We propose the Shaking Ratio metric to measure instability. We calculate it as the average motion of the central point between the frames transitions throughout the video, which is given by:
\begin{equation}
\label{eq:shakiness ratio}
\frac{1}{|\hat{V}|-1}\sum_{n = 1}^{|\hat{V}|-1} \frac{H(\hat{v}_n, \hat{v}_{n+1})}{d(v_n)},
\end{equation}
\noindent where $ \hat{v}_n $ is the $ n $-th frame in the output video, $ H $ computes the transition of the central point of $ \hat{v}_n $ when applying the estimated homography between $ \hat{v}_n $ and $ \hat{v}_{n+1} $, and $ d(\cdot) $ is the half of the frame diagonal. The lower is this value, the better it is. Whenever the homography cannot be estimated, we assign the value of the highest computed motion as a penalty.

\begin{table}[t!]
	\centering
	\caption{Average F\textsubscript{1} scores (higher is better) and Shaking Ratio (lower is better) for the output videos generated by all the compared methods in the three datasets (\%). Best results are in bold.}
	\label{tab:results}
	\setlength{\tabcolsep}{2.5pt}
	\footnotesize{
		\begin{tabular}{cclcccccccccc}
			\toprule
			\multirow{3}{*}{\thead{\rot[60][1em]{\footnotesize{\textbf{Dataset}}}}} & & \multirow{3}{*}{\thead{\rot[60][1em]{\footnotesize{\textbf{Method}}}}} & & \multicolumn{4}{c}{\thead{\footnotesize{\textbf{F\textsubscript{1} Score}}}} & & \thead{\multirow{3}{*}{\begin{tabular}{c}\footnotesize{\textbf{Shaking}}\\\footnotesize{\textbf{Ratio}}\end{tabular}}} \\ 
			& & & & \thead{\footnotesize{\textsc{Car}}} & \thead{\footnotesize{\textsc{Chair}}} & \thead{\footnotesize{\textsc{Comp.}}} & \thead{\footnotesize{\textsc{People}}}	& \thead{\footnotesize{\textsc{Tree}}}   &  &  \\  \cmidrule{5-9} 
			\multirow{4}{*}{\rotatebox[origin=c]{90}{\textit{UTE}}} 
			& & Unif.   		& & $ 09.6 $ & $ \mathbf{11.6} $ & $ 10.8 $ & $ 12.2 $ & $ 10.2 $ & $ 31.1 $ \\
			& &	MSH     		& & $ 10.2 $ & $ 10.5 $ & $ 08.3 $ & $ 12.7 $ & $ 11.1 $ & $ \mathbf{27.0} $ \\
			& & MIFF     		& & $ 10.4 $ & $ 10.3 $ & $ 06.1 $ & $ 13.9 $ & $ 11.6 $ & $ 47.1 $ \\
			& &\textbf{Ours} 	& & $ \mathbf{16.4} $  & $ 10.1 $ & $ \mathbf{23.6} $ & $ \mathbf{15.1} $ & $ \mathbf{18.1} $ &  $ 37.2 $ \\
			\\
			\multirow{4}{*}{\rotatebox[origin=c]{90}{\begin{tabular}{c}\textit{Semantic}\\\textit{Dataset}\end{tabular}}} 
			& & Unif.   		& & $ 12.9 $ & $ 07.3 $ & $ 06.9 $ & $ 12.2 $ & $ 15.2 $ & $ 11.0 $ \\
			& &	MSH     		& & $ 12.5 $ & $ 07.0 $ & $ 05.9 $ & $ 12.7 $ & $ 15.7 $ & $ \mathbf{04.4} $ \\
			& & MIFF     		& & $ 13.1 $ & $ \mathbf{09.1} $ & $ 07.4 $ & $ 13.9 $ & $ 13.6 $ & $ 08.9 $ \\
			& &\textbf{Ours} 	& & $ \mathbf{15.2} $  & $ 08.8 $ & $ \mathbf{07.5} $ & $ \mathbf{15.1} $ & $ \mathbf{18.5} $ & $ 10.1 $\\		
			\\
			\multirow{4}{*}{\rotatebox[origin=c]{90}{\begin{tabular}{c}\textit{Ego-}\\\textit{Sequences}\end{tabular}}} 
			& & Unif.   		& & $ 12.8 $ & $ 03.7 $ & $ 02.2 $ & $ 15.4 $ & $ 17.9 $ &  $ 12.0 $ \\
			& &	MSH     		& & $ 11.9 $ & $ 03.2 $ & $ 02.4 $ & $ 14.7 $ & $ 16.4 $ & $ \mathbf{04.7} $ \\
			& & MIFF     		& & $ 12.6 $ & $ 03.9 $ & $ 01.3 $ & $ \mathbf{17.2} $ & $ 15.4 $ & $ 08.2 $ \\
			& &\textbf{Ours} 	& & $ \mathbf{14.8} $  & $ \textbf{04.7} $ & $ \mathbf{04.4} $ & $ 16.4 $ & $ \mathbf{18.9} $ &  $ 08.2 $ 
			\\ 
			\bottomrule  
		\end{tabular}
	}
\end{table}
\paragraph{Evaluation model for simulated users.}
\label{sec:eval_model}
Aside from real Twitter users, we also evaluated our approach using virtual users. They were used for a more detailed performance assessment since we can control all aspects of their profiles.

We created five virtual Twitter users with interest in topics that are common in social networks and can be easily found in videos: Vehicles, Furniture, Technology, Human Interaction, and Nature. To represent each topic, we selected concepts based on the intersection of the SentiBank~\cite{Borth2013} and the PASCAL-Context dataset: \textsc{Car}, \textsc{Chair}, \textsc{Computer}, \textsc{People}, and \textsc{Tree}. Using the Twitter API, we collected over a week for \textit{tweets} containing these concepts in the hashtags (\#) and used them to feed a character-based LSTM network consisted of $ 3 $ recurrent layers of $ 512 $ units and dropout rate of $ 0.5 $. For each concept, we trained a different model to simulate \textit{tweets} written by each user. To prepare the data for training, we removed all special words, characters (\ie, @mentions, RTs, :, \$, !, \etc) and, to avoid bias in training, the hashtag terms were also removed. Finally, we have generated texts with the trained LSTM models to be used as input \textit{tweets} to our method.

\paragraph{Implementation Details.} For the \textit{word2vec} model, we used the parameters reported by Li~\etal~\cite{Li2017}. Thus, we refer the reader to their work for more details. Before clustering, we pruned out all words that were neither related to the Twitter vocabulary nor the VG dataset vocabulary since they rarely occur. This can be achieved by overlapping words in the VG vocabulary with Dataset7 and Dataset1~\cite{Li2017}. After this process, we ended up with $ {N = 936{,}225} $ embeddings. We tried several values for $ K $ (from $ 2^1 $ up to $ 2^{15} $). We used $ {K = 2^{13}} $ due to the insignificant reduction in the mean squared Euclidean distance from the word embeddings to their respective cluster centers when increasing the value of $ K $. We used the SentiStrength algorithm~\cite{Thelwall2010} to extract the positive sentences from the input text and optimized all $\lambda$ parameters ($\lambda_1$, $\lambda_2$, $\lambda_s$, $\lambda_i$, $\lambda_m$, and $\lambda_a$) using Particle Swarm Optimization~\cite{Silva2018jvci}. The desired speedup rate was set to $ {S = 10} $ for all experiments.

\subsection{Quantitative Results}

We report the average F\textsubscript{1} scores and Shaking Ratio values in Table~\ref{tab:results}. We used the texts from the simulated users and the videos of all datasets as input. Because only UTE contains human-annotated concepts, we used the nouns in the extracted sentences~\cite{Johnson2016} as concepts to validate the personalization in the EgoSequences and Semantic datasets. In the evaluation of the UTE Dataset, we replaced the concept \textsc{People} with \textsc{Men} since \textsc{People} is not in the annotations.

Regarding the personalization, the results show that our method outperforms all approaches in the majority of the concepts by a considerable margin, especially in the UTE Dataset, which contains human annotations. In our best results, our method gets average F\textsubscript{1} scores of $ 7.9 $ and $ 12.8 $ percentage points higher than the best competitors when using the text with \textit{tweets} about \textsc{Tree} and \textsc{Computer}, respectively. We accredit these results to our frame scoring approach, which is capable of using the context to infer the topics of interest for the user and assign higher scores for the scenes that exhibit the related concepts.

Notable exceptions are the experiments using \textsc{Chair}, in which the Uniform and MIFF approaches performed better than ours in UTE and Semantic datasets, respectively. However, we should note that the difference is marginal, being $ 1.5 $ percentage points in the UTE and $ 0.3 $ in the Semantic Dataset. We found that although \textsc{Chair} is present in the annotations of the UTE, this concept is not in the focus of attention in most frames since it is always coupled with other concepts that generally draw more attention (\eg, \textsc{Computer} and \textsc{Men}). Therefore, the output video emphasizes concepts found in the input texts other than \textsc{Chair}. We argue that the reason MIFF performed better in the Semantic Dataset is that in the walking and biking videos people are sat in chairs or benches and, since MIFF aims at emphasizing scenes with people, it produces a suitable frame selection.

With concerns to visual smoothness, the MSH produced output videos with the lowest Shaking Ratio values with at least $ 3.5 $ percentage points better than the best competitor. It is an expected result since MSH directly optimizes the smoothness in its frame selection. We also measured the speed-up deviation of the output videos. Our methodology achieved the best results for all cases, with an average value of $0.2$ facing $0.9$ and $1.4$ for MSH and MIFF, respectively.

\subsection{Evaluation by volunteers}

We used the output videos from EgoSequences generated by the semantic hyperlapse methods (MIFF and ours) to perform a survey. The volunteers were asked to watch a video prompted in a web page and answer: (i) select the most emphasized content (exhibited in a lower speed-up rate) and; (ii) evaluate the visual quality of the video. They were not informed which method generated the video.

As a preprocessing step to better extract meaningful results from the survey, we removed all videos with a low frequency of concepts since the emphasis applied by the speed-up change would not be perceptible, or the video would not change the playback speed at all. This step reduced the set of concepts to \textsc{Car}, \textsc{People}, and \textsc{Tree}. Also, to validate the data collected, we added two sets of placebo videos. For the first set, we manually selected $ 5 $ egocentric videos from GTEA Gaze+~\cite{Fathi2012} and created a semantic hyperlapse video for each one using the MIFF technique. We embedded the semantic information using the $ 24 $ classes from YOLO~\cite{Redmon2016} most related to food (apple, fork, spoon, \etc). The second set was composed of the output videos of the Uniform approach in the EgoSequences. The final collection of videos presented to the evaluators was composed of $ 40 $ fast-forward videos with an average length of $ 45 $ seconds. 

For the first question, we presented five mutually exclusive options representing the concepts: `Car', `People', `Tree', `Food', and `None of the above'. In the second question, the volunteer could rate the visual quality of the video as: `Very shaky', `Shaky', `Tolerable', `Smooth', and `Very smooth'.

\begin{table}[t!]
	\centering
	\caption{Average percentage of the selected concepts for question \#1 (\%). Higher values are in bold.}
	\label{tab:user_study}
	\footnotesize{
		\begin{tabular}{lcrrrrr}
			\toprule
			\thead{\textbf{Set}} & & \thead{\textsc{Car}} & \thead{\textsc{Tree}} & \thead{\textsc{People}} & \thead{\textsc{Food}} & \thead{None}  		\\  
			Ours+Car 		& & $ \mathbf{66.7} $ 	& $ 0.0 $ 			& $ 11.1 $ 			& $ 0.0 $ 			& $ 22.2 $ 			\\
			Ours+Tree 		& & $ 1.9 $ 			& $ \mathbf{42.3} $ & $ 25.0 $ 			& $ 1.9 $ 			& $ 28.9 $ 			\\
			Ours+People 	& & $ 9.5 $ 			& $ 2.4 $ 			& $ \mathbf{73.8} $ & $ 0.0 $ 			& $ 14.3 $ 			\\
			MIFF 			& & $ 8.9 $ 			& $ 6.7 $ 			& $ \mathbf{64.4} $ & $ 0.0 $ 			& $ 20.0 $ 			\\ \cdashlinelr{1-7}
			MIFF+Food 		& & $ 0.0 $ 			& $ 0.0 $ 			& $ 3.0 $ 			& $ \mathbf{87.9} $ & $ 9.1 $ 			\\
			Uniform 		& & $ 11.8 $ 			& $ 9.8 $ 			& $ 25.5 $ 			& $ 0.0 $ 			& $ \mathbf{52.9} $ \\
			\bottomrule 
		\end{tabular}
	}
\end{table}
\begin{figure}[t!]
	\centering
	\includegraphics[width=\columnwidth]{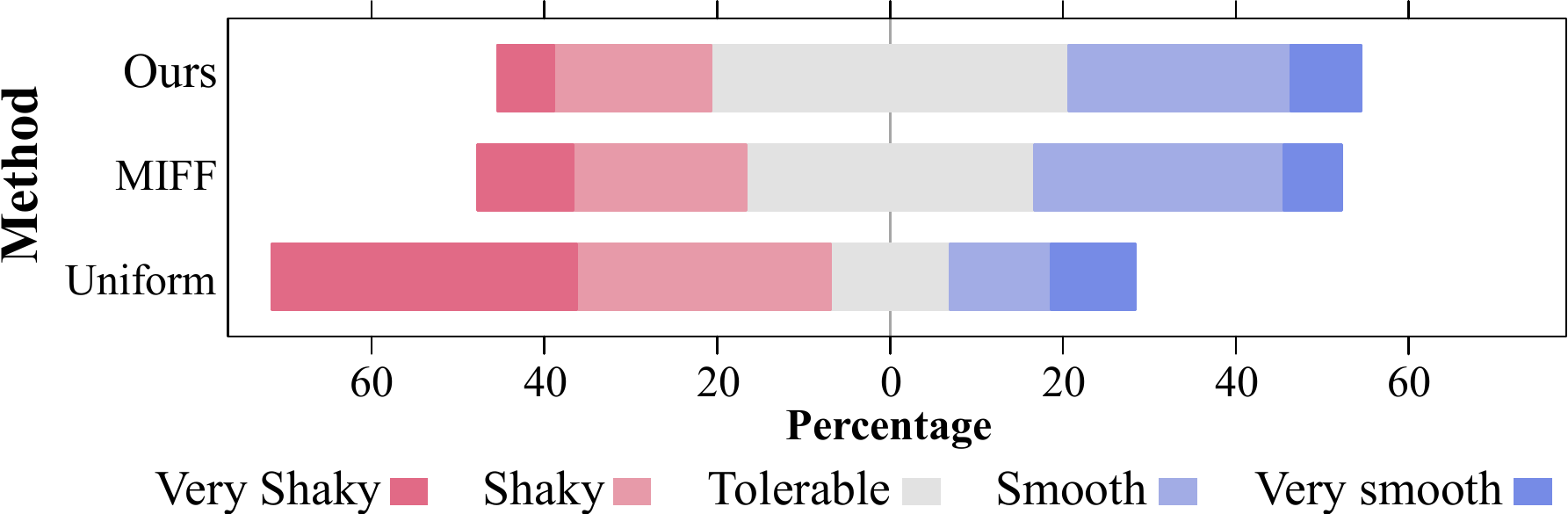}
	\caption{Likert scale plots for users' answers. Each bar represents the answers' distribution for the respective method. Bars are centered by the median value of the `Tolerable' answers.}
	\label{fig:likert}
\end{figure}
\begin{figure*}[t!]
	\centering
	\includegraphics[width=.975\linewidth]{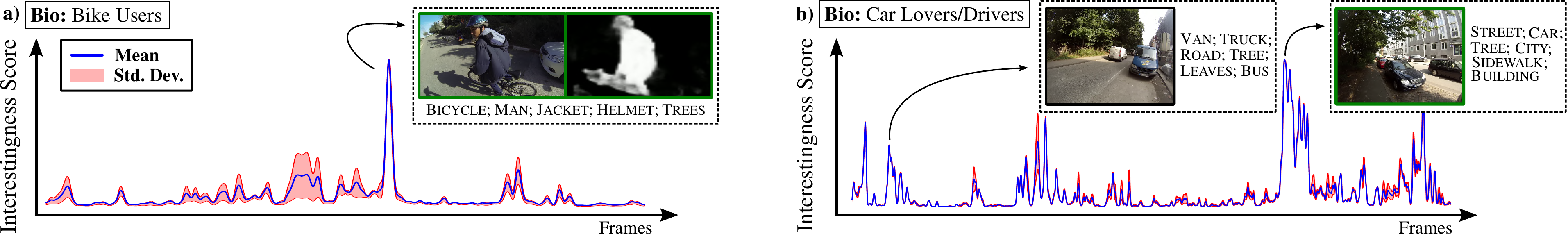}
	\caption{Mean and standard deviation of the Interestingness scores for the bike users in the video `Walking 3' (a), and the car lovers for the video `Bike 3' (b), both from EgoSequences. The green boxes present one of the most relevant frames for the users, according to our approach. Note that the extracted captions in the dashed boxes are closely related to the users' profile.}
	\label{fig:real_users_bike_and_car}
\end{figure*}

We collected $ 250 $ answers from $ 112 $ graduate or undergraduate students from the computer science department. Table~\ref{tab:user_study} presents the average percentage of the selected concepts for each set of videos. The sets with the pattern `Ours+$<$Concept$>$' represent the output videos of our technique when using as input the sentences generated by the LSTM model that \textit{tweets} about the $<$Concept$>$. The set `MIFF' represents the videos generated by using the approach of Silva~\etal~\cite{Silva2018jvci} with either face or pedestrian, as reported in their paper. Below the dashed line are the placebo sets, where the set `MIFF+Food' represents the videos from the GTEA Gaze+ dataset, and `Uniform' represents the videos generated by the Uniform approach.

As expected for the sets of placebo videos, the majority of the evaluators selected `Food' for most videos in the `MIFF+Food' set and `None of the above' for the videos in the `Uniform' set, validating our survey. However, it should be noted that for the `Uniform' set, $ 25.5\% $ of the evaluators considered \textsc{People} as being emphasized in the videos. We argue that \textsc{People} is a concept of common interest, and the evaluators might have been more attentive when people appear, leading them to this conclusion. In fact, \textsc{People} is the only concept that was selected at least once, regardless of the presented set.

Most of the volunteers ($ 73.8\% $) selected `People' after watching the videos generated by our technique when using the concept \textsc{People}, while $ 64.44\% $ selected `People' after watching the ones generated by MIFF. This result demonstrates that our approach was capable of selecting as many relevant frames as MIFF when using this concept. Furthermore, our algorithm achieved the correct selection in $ 66.7\% $ and $ 42.3\% $ of the answers for the sets `Ours+Car' and `Ours+Tree', respectively, corresponding to the major part of the answers. Therefore, we can draw the following observation: our approach could personalize semantics to match the users' interest and also generate better results when compared to MIFF.

We report the results for the answers to the second question in the Likert scale plot portrayed in Figure~\ref{fig:likert}. Each bar represents the answers' distribution for the respective method, centered by the median value of the `Tolerable' answers. Users considered the Uniform method as being the least pleasant to watch. The negative answers (`Shaky' and `Very shaky') were $64.7\%$ of the answers for the Uniform method. Both our method and MIFF received a similar evaluation regarding their visual quality, which was on average `Tolerable'. Although our approach has the personalized semantics constraint, its non-negative answers ($75.2\%$), compared to MIFF's ($68.9\%$), indicate that we do not compromise the perceived stability.

\subsection{Results with Twitter Users}

In this section, we evaluate our frame scoring approach using real users as input. We manually selected active users on \textit{Twitter} who have explicitly indicated topics of interest in their bio (self-written short biography in their profile), including sentences such as ``love $<$concept$>$'' or ``$<$concept$>$ enthusiast''. For each concept, we selected $ 5 $ users and collected their last $ 3{,}000 $ \textit{tweets}, when available. Similar to the evaluation model for simulated users, we pre-filtered the input texts and ended up with $ {\sim1{,}000} $ \textit{tweets} for each user after sentiment analysis. We selected users with the following profiles: cyclists; car lovers/drivers; and gardeners. Moreover, we selected representative videos from the datasets containing a wide range of concepts, which gives us a valuable discussion of the components of our methodology.

Fig.~\ref{fig:real_users_bike_and_car} depicts the mean (blue line) and standard deviation (red shaded region) of the semantic scores assigned by our approach for the users in each video. A high mean value in combination with a low standard deviation indicates a video segment of simultaneous interest among the users. A sizable shaded region indicates divergence among users.

Fig.~\ref{fig:real_users_bike_and_car}-a presents the interestingness scores for the cyclists in the video `Walking 3' from EgoSequences. The picture in the green box shows one of the frames with the highest score (left), the saliency map (right), and the extracted concepts (bottom). Note that one of the extracted concepts was \textsc{Bicycle}, which matches the users' profiles. Although the cyclists have diverse interests over the video, the frame of high mutual interest presents a man riding a bike, which is a singular moment in this video, reinforcing the importance of using the uniqueness score. It is noteworthy that no visual features are extracted from the users' profiles by any component of our methodology. The users' interests are inferred from raw texts in their social network profiles.

Frame scoring results for the car lovers/drivers in the video `Bike 3' from the EgoSequences are presented in Fig.~\ref{fig:real_users_bike_and_car}-b. The rightmost dashed box includes the frame of highest interest according to our approach and its respective extracted concepts. Because the users \textit{tweet} positively about \textsc{Cars}, their BoT present higher activations in the vehicles cluster, which allows assigning reasonable scores to frames that the concept \textsc{Car} is not present, but their semantic-related concepts are (\ie, \textsc{Van}, \textsc{Taxi}, \textsc{Buses}, \etc). We illustrate this case in the black dashed box on the left.

\section{Discussion and future work}
\label{sec:conclusions}

In this paper, we presented a novel approach capable of creating personalized hyperlapse videos, extracting meaningful moments for a watcher according to her/his activity in social networks. Our methodology mines sentences from the user's social network to infer topics of interest, presenting average F\textsubscript{1} scores of up to $ 12.8 $ percentage points higher than the best competitor. We also presented a user study to validate the effectiveness of our methodology \wrt personalization and stabilization aspects by presenting a personalized hyperlapse regarding a specific topic. User-perceived topics matched the topics used to generate the accelerated video in most cases, about $61\%$ on average.

Despite the promising results, our method may fail to emphasize the relevant content in a video if semantically related concepts lie on distinct clusters, \eg, a video segment containing \textsc{Trucks} could be not emphasized for a user who posts about \textsc{Cars} if \textsc{Trucks} and \textsc{Cars} are from different clusters. Also, in applications that neither the content novelty nor the visual saliency matter, a limitation arises since objects can rely on an image patch of low visual saliency, or the TF-IDF is low due to its recurrence. This results in a video segment being assigned a lower score, even containing objects that match an interesting concept for the user.

In future works, we pursue to explore different types of media (\eg, images, videos, and sound) shared by the users in social networks. We believe a multi-modal approach might be a promising research direction for modeling user behavior and better refining semantics in FPVs.

\paragraph{Acknowledgments.} We thank the agencies CAPES, CNPq, and FAPEMIG for funding different parts of this work. We also thank NVIDIA Corporation for the donation of a Titan XP GPU used in this research.

{\small
\bibliographystyle{ieee}
\bibliography{wacv_ramos2020}

\begin{thebibliography}{10}\itemsep=-1pt

\bibitem{Bhattacharya2014}
P.~Bhattacharya, M.~B. Zafar, N.~Ganguly, S.~Ghosh, and K.~P. Gummadi.
\newblock Inferring user interests in the twitter social network.
\newblock In {\em ACM Conference on Recommender Systems}, RecSys '14, pages
  357--360, New York, NY, USA, 2014. ACM.

\bibitem{Borth2013}
D.~Borth, T.~Chen, R.~Ji, and S.-F. Chang.
\newblock Sentibank: Large-scale ontology and classifiers for detecting
  sentiment and emotions in visual content.
\newblock In {\em ACM International Conference on Multimedia}, MM '13, pages
  459--460, New York, NY, USA, 2013. ACM.

\bibitem{Donahue2017}
J.~{Donahue}, L.~A. {Hendricks}, M.~{Rohrbach}, S.~{Venugopalan},
  S.~{Guadarrama}, K.~{Saenko}, and T.~{Darrell}.
\newblock Long-term recurrent convolutional networks for visual recognition and
  description.
\newblock {\em {IEEE} Transactions on Pattern Analysis and Machine Intelligence
  ({TPAMI})}, 39(4):677--691, 2017.

\bibitem{Dong2018}
J.~Dong, X.~Li, and C.~G.~M. Snoek.
\newblock Predicting visual features from text for image and video caption
  retrieval.
\newblock {\em {IEEE} Transactions on Multimedia ({TMM})}, 20(12):3377--3388,
  Dec 2018.

\bibitem{Fang2015}
H.~{Fang}, S.~{Gupta}, F.~N. {Iandola}, R.~K. {Srivastava}, L.~{Deng},
  P.~{Dollár}, J.~{Gao}, X.~{He}, M.~{Mitchell}, J.~C. {Platt}, C.~L.
  {Zitnick}, and G.~{Zweig}.
\newblock From captions to visual concepts and back.
\newblock In {\em {IEEE} Conference on Computer Vision and Pattern Recognition
  ({CVPR})}, pages 1473--1482, 2015.

\bibitem{Fathi2012}
A.~Fathi, Y.~Li, and J.~M. Rehg.
\newblock Learning to recognize daily actions using gaze.
\newblock In {\em European Conference on Computer Vision ({ECCV})}, ECCV'12,
  pages 314--327, Berlin, Heidelberg, 2012. Springer-Verlag.

\bibitem{FeiFei2005}
L.~Fei-Fei and P.~Perona.
\newblock A bayesian hierarchical model for learning natural scene categories.
\newblock In {\em {IEEE} Conference on Computer Vision and Pattern Recognition
  ({CVPR})}, volume~2, pages 524--531 vol. 2, June 2005.

\bibitem{Halperin2017}
T.~{Halperin}, Y.~{Poleg}, C.~{Arora}, and S.~{Peleg}.
\newblock Egosampling: Wide view hyperlapse from egocentric videos.
\newblock {\em {IEEE} Transactions on Circuits and Systems for Video Technology
  ({TCSVT})}, 28(5):1248--1259, May 2018.

\bibitem{Higuchi2017}
K.~Higuchi, R.~Yonetani, and Y.~Sato.
\newblock Egoscanning: Quickly scanning first-person videos with egocentric
  elastic timelines.
\newblock In {\em The Conference on Human Factors in Computing Systems (CHI)},
  CHI '17, pages 6536--6546, New York, NY, USA, 2017. ACM.

\bibitem{Johnson2016}
J.~{Johnson}, A.~{Karpathy}, and L.~{Fei-Fei}.
\newblock Densecap: Fully convolutional localization networks for dense
  captioning.
\newblock In {\em {IEEE} Conference on Computer Vision and Pattern Recognition
  ({CVPR})}, pages 4565--4574, June 2016.

\bibitem{Joshi2015}
N.~Joshi, W.~Kienzle, M.~Toelle, M.~Uyttendaele, and M.~F. Cohen.
\newblock Real-time hyperlapse creation via optimal frame selection.
\newblock {\em ACM Transactions on Graphics (TOG)}, 34(4):63:1--63:9, July
  2015.

\bibitem{Karpathy2017}
A.~{Karpathy} and L.~{Fei-Fei}.
\newblock Deep visual-semantic alignments for generating image descriptions.
\newblock {\em {IEEE} Transactions on Pattern Analysis and Machine Intelligence
  ({TPAMI})}, 39(4):664--676, 2017.

\bibitem{Karpenko2014}
A.~Karpenko.
\newblock The technology behind hyperlapse from instagram.
\newblock http://instagram-engineering.tumblr.com/post/95922900787/hyperlapse,
  Aug. 2014.
\newblock Accessed 25 March 2019.

\bibitem{Kopf2014}
J.~Kopf, M.~F. Cohen, and R.~Szeliski.
\newblock First-person hyper-lapse videos.
\newblock {\em ACM Transactions on Graphics (TOG)}, 33(4):78:1--78:10, July
  2014.

\bibitem{Kottur2016}
S.~Kottur, R.~Vedantam, J.~M.~F. Moura, and D.~Parikh.
\newblock Visualword2vec (vis-w2v): Learning visually grounded word embeddings
  using abstract scenes.
\newblock In {\em {IEEE} Conference on Computer Vision and Pattern Recognition
  ({CVPR})}, pages 4985--4994, June 2016.

\bibitem{Krishna2017}
R.~{Krishna}, Y.~{Zhu}, O.~{Groth}, J.~{Johnson}, K.~{Hata}, J.~{Kravitz},
  S.~{Chen}, Y.~{Kalantidis}, L.-J. {Li}, D.~A. {Shamma}, M.~S. {Bernstein},
  and L.~{Fei-Fei}.
\newblock Visual genome: Connecting language and vision using crowdsourced
  dense image annotations.
\newblock {\em International Journal of Computer (IJC)}, 123(1):32--73, 2017.

\bibitem{Lai2017}
W.~{Lai}, Y.~{Huang}, N.~{Joshi}, C.~{Buehler}, M.~{Yang}, and S.~B. {Kang}.
\newblock Semantic-driven generation of hyperlapse from 360 degree video.
\newblock {\em IEEE Transactions on Visualization and Computer Graphics},
  24(9):2610--2621, Sep. 2018.

\bibitem{Lan2018}
S.~{Lan}, R.~{Panda}, Q.~{Zhu}, and A.~K. {Roy-Chowdhury}.
\newblock Ffnet: Video fast-forwarding via reinforcement learning.
\newblock In {\em {IEEE} Conference on Computer Vision and Pattern Recognition
  ({CVPR})}, pages 6771--6780, June 2018.

\bibitem{Lee2012}
Y.~J. Lee, J.~Ghosh, and K.~Grauman.
\newblock Discovering important people and objects for egocentric video
  summarization.
\newblock In {\em {IEEE} Conference on Computer Vision and Pattern Recognition
  ({CVPR})}, pages 1346--1353, June 2012.

\bibitem{Li2017}
Q.~Li, S.~Shah, X.~Liu, and A.~Nourbakhsh.
\newblock Data sets: Word embeddings learned from tweets and general data.
\newblock In {\em AAAI Conference on Web and Social Media}, 2017.

\bibitem{Michelson2010}
M.~Michelson and S.~A. Macskassy.
\newblock Discovering users' topics of interest on twitter: A first look.
\newblock In {\em Workshop on Analytics for Noisy Unstructured Text Data}, AND
  '10, pages 73--80, New York, NY, USA, 2010. ACM.

\bibitem{Mikolov2013}
T.~Mikolov, I.~Sutskever, K.~Chen, G.~S. Corrado, and J.~Dean.
\newblock Distributed representations of words and phrases and their
  compositionality.
\newblock In {\em Advances in Neural Information Processing Systems}, pages
  3111--3119. Curran Associates, Inc., 2013.

\bibitem{Mottaghi2014}
R.~Mottaghi, X.~Chen, X.~Liu, N.~G. Cho, S.~W. Lee, S.~Fidler, R.~Urtasun, and
  A.~Yuille.
\newblock The role of context for object detection and semantic segmentation in
  the wild.
\newblock In {\em {IEEE} Conference on Computer Vision and Pattern Recognition
  ({CVPR})}, pages 891--898, June 2014.

\bibitem{Passalis2018}
N.~Passalis and A.~Tefas.
\newblock Learning bag-of-embedded-words representations for textual
  information retrieval.
\newblock {\em Pattern Recognition}, 81:254 -- 267, 2018.

\bibitem{Plummer2017}
B.~A. {Plummer}, M.~{Brown}, and S.~{Lazebnik}.
\newblock Enhancing video summarization via vision-language embedding.
\newblock In {\em {IEEE} Conference on Computer Vision and Pattern Recognition
  ({CVPR})}, pages 5781--5789, 2017.

\bibitem{Poleg2015}
Y.~Poleg, T.~Halperin, C.~Arora, and S.~Peleg.
\newblock Egosampling: Fast-forward and stereo for egocentric videos.
\newblock In {\em {IEEE} Conference on Computer Vision and Pattern Recognition
  ({CVPR})}, pages 4768--4776, June 2015.

\bibitem{Ramos2016}
W.~L.~S. Ramos, M.~M. Silva, M.~F.~M. Campos, and E.~R. Nascimento.
\newblock Fast-forward video based on semantic extraction.
\newblock In {\em {IEEE} International Conference on Image Processing
  ({ICIP})}, pages 3334--3338, Sept 2016.

\bibitem{Rani2018}
P.~Rani, A.~Jangid, V.~P. Namboodiri, and K.~S. Venkatesh.
\newblock Visual odometry based omni-directional hyperlapse.
\newblock In {\em National Conference on Computer Vision, Pattern Recognition,
  Image Processing, and Graphics}, pages 3--13, Singapore, 2018. Springer
  Singapore.

\bibitem{Redmon2016}
J.~{Redmon}, S.~K. {Divvala}, R.~B. {Girshick}, and A.~{Farhadi}.
\newblock You only look once: Unified, real-time object detection.
\newblock In {\em {IEEE} Conference on Computer Vision and Pattern Recognition
  ({CVPR})}, pages 779--788, 2016.

\bibitem{Sazbon2004}
D.~Sazbon, H.~Rotstein, and E.~Rivlin.
\newblock Finding the focus of expansion and estimating range using optical
  flow images and a matched filter.
\newblock {\em Machine Vision and Applications (MVA)}, 15(4):229--236, 2004.

\bibitem{Sharghi2016}
A.~{Sharghi}, B.~{Gong}, and M.~{Shah}.
\newblock Query-focused extractive video summarization.
\newblock {\em European Conference on Computer Vision ({ECCV})}, pages 3--19,
  2016.

\bibitem{Sharghi2017}
A.~{Sharghi}, J.~S. {Laurel}, and B.~{Gong}.
\newblock Query-focused video summarization: Dataset, evaluation, and a memory
  network based approach.
\newblock In {\em {IEEE} Conference on Computer Vision and Pattern Recognition
  ({CVPR})}, pages 4788--4797, 2017.

\bibitem{Silva2018jvci}
M.~M. Silva, W.~L.~S. Ramos, F.~C. Chamone, J.~P.~K. Ferreira, M.~F.~M. Campos,
  and E.~R. Nascimento.
\newblock Making a long story short: A multi-importance fast-forwarding
  egocentric videos with the emphasis on relevant objects.
\newblock {\em Journal of Visual Communication and Image Representation
  (JVCI)}, 53:55 – 64, 2018.

\bibitem{Silva2016}
M.~M. Silva, W.~L.~S. Ramos, J.~P.~K. Ferreira, M.~F.~M. Campos, and E.~R.
  Nascimento.
\newblock Towards semantic fast-forward and stabilized egocentric videos.
\newblock In {\em European Conference on Computer Vision Workshop ({ECCVW})},
  pages 557--571, Amsterdam, NL, October 2016. Springer International
  Publishing.

\bibitem{Silva2018cvpr}
M.~M. Silva, W.~L.~S. Ramos, J.~P.~K. Ferreira, F.~C. Chamone, M.~F.~M. Campos,
  and E.~R. Nascimento.
\newblock A weighted sparse sampling and smoothing frame transition approach
  for semantic fast-forward first-person videos.
\newblock In {\em {IEEE} Conference on Computer Vision and Pattern Recognition
  ({CVPR})}, pages 2383--2392, Salt Lake City, USA, Jun. 2018.

\bibitem{Thelwall2010}
M.~{Thelwall}, K.~{Buckley}, G.~{Paltoglou}, D.~{Cai}, and A.~{Kappas}.
\newblock Sentiment strength detection in short informal text.
\newblock {\em Journal of the Association for Information Science and
  Technology}, 61(12):2544--2558, 2010.

\bibitem{Varini2017}
P.~Varini, G.~Serra, and R.~Cucchiara.
\newblock Personalized egocentric video summarization of cultural tour on user
  preferences input.
\newblock {\em {IEEE} Transactions on Multimedia ({TMM})}, 19(12):2832--2845,
  Dec 2017.

\bibitem{Venugopalan2015}
S.~Venugopalan, M.~Rohrbach, J.~Donahue, R.~J. Mooney, T.~Darrell, and
  K.~Saenko.
\newblock Sequence to sequence -- video to text.
\newblock In {\em {IEEE} International Conference on Computer Vision ({ICCV})},
  pages 4534--4542, 2015.

\bibitem{Vinyals2017}
O.~Vinyals, A.~Toshev, S.~Bengio, and D.~Erhan.
\newblock Show and tell: Lessons learned from the 2015 mscoco image captioning
  challenge.
\newblock {\em {IEEE} Transactions on Pattern Analysis and Machine Intelligence
  ({TPAMI})}, 39(4):652--663, April 2017.

\bibitem{WangM2018}
M.~Wang, J.~Liang, S.~Zhang, S.~Lu, A.~Shamir, and S.~Hu.
\newblock Hyper-lapse from multiple spatially-overlapping videos.
\newblock {\em IEEE Transactions on Image Processing}, 27(4):1735--1747, April
  2018.

\bibitem{WangW2018}
W.~Wang, J.~Shen, and L.~Shao.
\newblock Video salient object detection via fully convolutional networks.
\newblock {\em IEEE Transactions on Image Processing}, 27(1):38--49, Jan 2018.

\bibitem{Yao2016}
T.~{Yao}, T.~{Mei}, and Y.~{Rui}.
\newblock Highlight detection with pairwise deep ranking for first-person video
  summarization.
\newblock In {\em {IEEE} Conference on Computer Vision and Pattern Recognition
  ({CVPR})}, pages 982--990, June 2016.

\end{thebibliography}
}

\end{document}